\pdfoutput=1
\RequirePackage{fix-cm}
\documentclass{article}

\PassOptionsToPackage{numbers, compress}{natbib}
\usepackage[preprint]{neurips_2026}
\usepackage[utf8]{inputenc}
\usepackage[T1]{fontenc}
\usepackage[bottom]{footmisc}
\makeatletter
\renewcommand{\@notice}{\enlargethispage{2\baselineskip}}
\makeatother
\usepackage{xcolor}
\definecolor{r2sblue}{HTML}{2563EB}
\definecolor{r2sindigo}{HTML}{4F46E5}
\definecolor{r2sviolet}{HTML}{7C3AED}
\definecolor{r2sink}{HTML}{1E293B}
\definecolor{r2sgray}{HTML}{64748B}
\definecolor{r2spanel}{HTML}{F5F7FF}
\usepackage[colorlinks=true, linkcolor=r2sindigo, citecolor=r2sviolet, urlcolor=r2sblue]{hyperref}
\usepackage{url}
\usepackage{booktabs}
\usepackage{tabularx}
\usepackage{colortbl}
\newcolumntype{Y}{>{\raggedright\arraybackslash}X}
\usepackage{amsmath}
\usepackage{amssymb}
\usepackage{microtype}
\usepackage{graphicx}
\usepackage{enumitem}
\usepackage{fontawesome5}
\usepackage{marvosym}
\usepackage{caption}
\DeclareCaptionType{listing}[Listing][List of Listings]
\usepackage[most]{tcolorbox}
\usepackage{titlesec}
\titleformat{\section}{\large\bfseries}{\thesection}{0.7em}{}
\titleformat{\subsection}{\normalsize\bfseries}{\thesubsection}{0.7em}{}
\titleformat{\paragraph}[runin]{\normalsize\bfseries}{}{0pt}{}
\titlespacing*{\section}{0pt}{1.6ex plus .5ex minus .2ex}{0.9ex plus .2ex}
\titlespacing*{\subsection}{0pt}{1.2ex plus .4ex minus .2ex}{0.6ex plus .2ex}
\titlespacing*{\paragraph}{0pt}{0.9ex plus .3ex minus .2ex}{0.7em}

\makeatletter
\renewcommand{\@toptitlebar}{%
  \hbox{\tikz\shade[left color=r2sblue, middle color=r2sindigo, right color=r2sviolet]
    (0,0) rectangle (\textwidth,3pt);}%
  \vskip 0.1in
  \vskip -\parskip
}
\renewcommand{\@bottomtitlebar}{%
  \vskip 0.24in
  \vskip -\parskip
  {\color{r2sindigo!45}\hrule height 0.8\p@}%
  \vskip 0.02in
}
\makeatother

\newtcolorbox{r2sabstract}{enhanced, colback=white, colframe=r2sindigo!50, boxrule=0.5pt,
  arc=2.5mm, left=7.5mm, right=7.5mm, top=3.8mm, bottom=4.4mm, before skip=4pt, after skip=9pt}
\renewenvironment{abstract}{%
  \begin{r2sabstract}
  \centerline{\fontfamily{Inter-LF}\fontseries{b}\fontsize{12}{14}\selectfont\color{r2sindigo}Abstract}%
  \vspace{6pt}%
  \fontfamily{SourceSerifFour-TLF}\fontsize{9.4}{12.4}\selectfont\color{r2sink}%
}{\end{r2sabstract}}

\usepackage{fancyhdr}
\AtBeginDocument{\setlength{\headwidth}{\textwidth}}

\newcommand{\lab}{Real2SimLab}
\newcommand{\cmd}[1]{\texttt{#1}}

\lstdefinelanguage{toml}{
  morekeywords={true, false},
  sensitive=true,
  morecomment=[l]{\#},
  morestring=[b]",
}
\newtcblisting{r2slisting}[1][]{
  listing only, enhanced, colback=r2spanel, colframe=r2sindigo!35, arc=2mm, boxrule=0.6pt,
  left=2mm, right=2mm, top=0.6mm, bottom=0.6mm,
  listing options={language=toml, basicstyle=\ttfamily\scriptsize, columns=fullflexible,
    keepspaces=true, commentstyle=\color{r2sgray}, stringstyle=\color{r2sviolet},
    keywordstyle=\color{r2sblue}},
  #1
}

\definecolor{r2sbrand}{HTML}{2E1065}
\definecolor{r2slabl}{HTML}{6D28D9}
\definecolor{r2slaba}{HTML}{2563EB}
\definecolor{r2slabb}{HTML}{0EA5C9}
\newcommand{\labname}{\textcolor{r2sbrand}{Real2Sim}\textcolor{r2slabl}{L}\textcolor{r2slaba}{a}\textcolor{r2slabb}{b}}

\title{\raisebox{-0.24\height}{\includegraphics[height=1.3em]{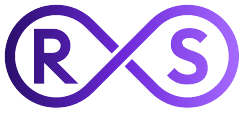}}\hspace{0.45em}%
\labname: A Unified Platform\\ for Real-to-Simulation Transfer}

\author{%
Yongchang Zhang$^{\star,\dagger,1}$,
Jie Huang$^{\star,\dagger,1}$,
Yunhe Ni$^{\star,1}$,
Ningyu Zhou$^{3}$,\\[2.5pt]
\textbf{
Guangyu Zhao$^{1}$,
Ruixun Liu$^{1}$,
Rui Huang$^{4}$,
Yidi Wu$^{3}$,
TianZe Zhang$^{1}$,}\\[2.5pt]
\textbf{
Zixiang Jiang$^{2}$,
Xiang Li$^{2}$,
Kevin Shen$^{2}$,
Yiwu Zhong$^{1\textrm{\Letter}}$
} 
\\[4.5pt]
$^{1}$ State Key Laboratory of General AI, School of Intelligence Science and Technology, PKU \\[2pt]
$^{2}$ Ngoro Robotics \quad
$^{3}$ School of Physics, PKU \quad 
$^{4}$ The University of Hong Kong \\[2pt]
$^{\star}$ Equal contribution \quad
$^{\dagger}$ Co-project leaders \quad
$\textrm{\Letter}$ Corresponding author
}

\begin{document}

\maketitle

\vspace{-2.5em}
\begin{center}
  \small
  \textcolor{r2sink}{\faGithub}~\href{https://github.com/Real2SimHub/Real2SimLab}{\texttt{Real2SimLab}}
  \qquad
  \textcolor{r2sink}{\faGithub}~\href{https://github.com/Real2SimHub/Awesome-Real2Sim}{\texttt{Awesome-Real2Sim}}
  \qquad
  \textcolor{r2sink}{\faGlobe}~\href{https://real2simhub.github.io/}{Real2SimHub}
\end{center}
\begin{abstract}
Real-to-simulation transfer, which builds simulations from real-world observations, has long been studied in research and industry.
Recently, coding agents have become capable of generating simulation programs from video recordings, and several benchmarks have been proposed to evaluate them.
However, each individual benchmark defines its own inputs, deliverables, and scoring pipeline. This leads to $N \times M$ integrations when evaluating $N$ methods on $M$ benchmarks.
To this end, we present \textbf{\lab}, a unified platform for real-to-simulation transfer.
It evaluates a method by running its written program and scoring the output simulation with the original scorer of each benchmark.
Moreover, \lab{} provides a pipeline to create training data and reward annotations for real-to-simulation methods. 
Our pipeline carefully avoids data contamination by filtering out the video recordings that exist in benchmarks. 
For each video recording, a program is first produced, and then our pipeline computes its reward with the same code that \lab{} uses for evaluation.
So far, it supports 47~coding agents and 5~non-agent methods on 6~benchmarks. 
Both the platform and \textbf{Awesome-Real2Sim}, our research map of nearly 300 real-to-simulation papers, are \textbf{open source}, and we invite the \textbf{Real2SimHub community} to build them together.
\end{abstract}

\begin{figure}[h]
  \centering
  \includegraphics[width=\linewidth]{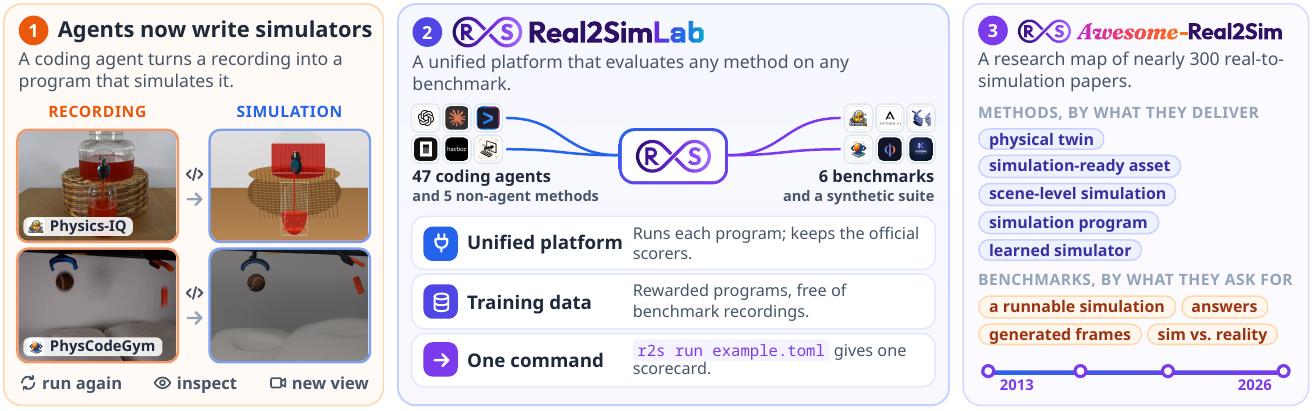}
  \caption{\textbf{Overview of this work.}
  (1)~The new paradigm: a coding agent turns a recording of a real event into a program that simulates it.
  Every simulation frame here is rendered by re-running a program that the Codex agent wrote for the input video recording.%
  (2)~\lab{} runs the program that a method writes, scores its output with the scorer of the benchmark, and creates training data and reward annotations.
  (3)~Awesome-Real2Sim organizes nearly 300 real-to-simulation papers according to what they deliver or ask for.}
  \label{fig:overview}
\end{figure}

\section{Introduction}
\label{sec:intro}

Real-to-simulation transfer has long been an active topic in both research and industry.
It takes real-world observations, such as images, videos, and 3D scans, to build simulations in which real objects, scenes, and events can be reproduced, analyzed, and reused.
This field spans computer vision, graphics, and robotics, and most of its methods are designed for one kind of input and one kind of output.
For example, some methods fit physical parameters, such as mass and friction, to videos of objects~\cite{wu2015galileo,wu2017vda,jatavallabhula2021gradsim,li2023pacnerf,jiang2025phystwin}, some assign physical properties to reconstructed 3D scenes~\cite{xie2024physgaussian,zhang2024physdreamer}, and others build sensor simulators from real scans~\cite{manivasagam2020lidarsim} or simulated training environments for robots from real rooms~\cite{torne2024rialto,dai2024acdc}.
Each of these methods is a dedicated pipeline, which requires a new pipeline for a new kind of scene.

Recently, rapid progress in foundation models, such as Claude Fable~5.2~\cite{anthropic2026fable} and GPT-6~\cite{openai2026gpt6}, has made coding agents capable of writing and debugging large programs in an evolving way.
This capability opens a new paradigm for real-to-simulation transfer (Figure~\ref{fig:overview}, panel~1): instead of building a dedicated pipeline for each kind of scene, one can ask a coding agent to write a simulation program directly from a video recording~\cite{cherian2024llmphy,chen2026agentic}.
For example, given a video of a red drink being poured into a glass, an agent can write a program that sets up the glass and the liquid in a physics simulator and reproduces the pour.
While the video shows the event only once and from one viewpoint, the program can replay the event, render it from new viewpoints, and expose the state of every object at every moment.
The same agent can attempt many kinds of scenes, offering a promising step toward building simulations in an automatic way.
In this paper, we focus on such a setting, in which a method, such as a coding agent, receives a video recording and delivers a simulation program that can be run.

\begin{figure}[h]
  \centering
  \includegraphics[width=\linewidth]{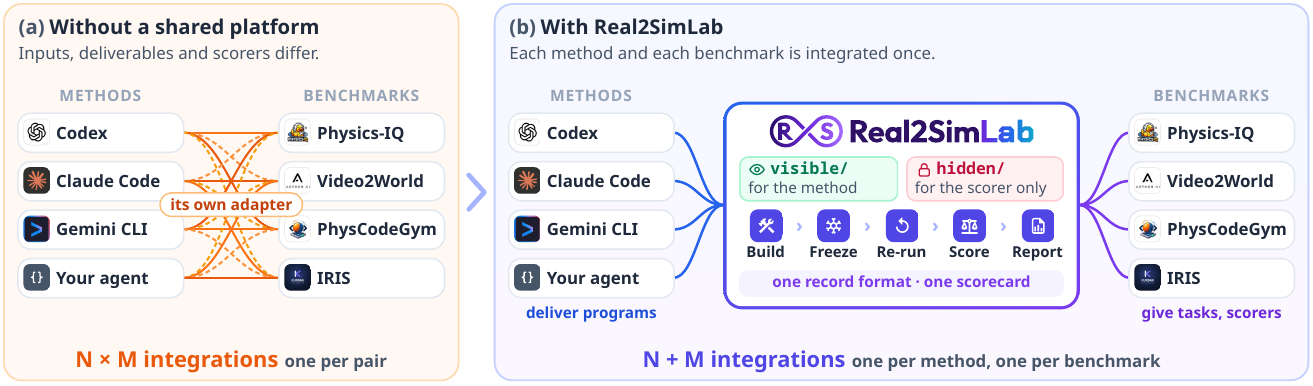}
  \caption{\textbf{Why a shared platform helps.}
  (a)~Without a shared platform, benchmarks differ in what they provide to a method, what they ask the method to deliver, and how they score the output. Every pair of a method and a benchmark thus requires its own adapter to run.
  (b)~\lab{} makes the delivered program into the interface between methods and benchmarks. So each method and each benchmark is integrated only once.}
  \label{fig:integrations}
\end{figure}

To measure the progress of the methods, several benchmarks have been proposed~\cite{motamed2026physics,khanbayov2026iris,shen20264dcodebench,shao2026physcodegym,tang2026video2world,kundu2026physimcode}.
They contribute valuable recordings, carefully designed tasks, and metrics.
However, each benchmark defines its own inputs, deliverables, and scoring pipeline.
For example, Physics-IQ~\cite{motamed2026physics} gives a method the first three seconds of a real video and scores the video frames that the method produces for the next five seconds, whereas Video2World~\cite{tang2026video2world} gives a whole robot manipulation video and asks for a simulation package that contains the reconstructed scene and the actions of the robot.
Other benchmarks ask for other types of deliverables: 4DCodeBench~\cite{shen20264dcodebench} asks for a script that builds a 4D world, that is, a 3D scene that changes over time, and PhysCodeGym~\cite{shao2026physcodegym} asks for a Blender scene that reproduces a reference video.
Each benchmark then scores its deliverable with its own code, in its own software environment.
As a result, a method built for one benchmark cannot be evaluated on another without a dedicated adapter, and evaluating $N$ methods on $M$ benchmarks requires $N \times M$ integrations (Figure~\ref{fig:integrations}a).
This cost discourages the evaluation of a method on more than a few benchmarks, and it is also hard to compare the results from different papers.

To this end, we present \textbf{\lab}, a unified platform for real-to-simulation transfer (Figure~\ref{fig:overview}, panel~2).
It first runs the program that a method writes and scores the output simulation with the original scorer of a benchmark.
Since every method delivers a program and every benchmark scores what a program outputs, the program becomes a common interface between the two sides (Figure~\ref{fig:integrations}b).
On the method side, each method is connected once, through a method-wise adapter that passes a task to the method and collects its program.
On the benchmark side, each benchmark is connected once, through a benchmark-wise adapter that converts its data into tasks of one shared format and keeps the official scorer.
For example, on Physics-IQ, a method now delivers a program that writes the requested frames, and the official scorer evaluates these frames as the original benchmark.
Evaluating $N$ methods on $M$ benchmarks then requires only $N + M$ integrations instead of $N \times M$.

Moreover, \lab{} provides a pipeline that creates training data with reward annotations for real-to-simulation methods.
These training data should be free of contamination. However, data contamination oftentimes happens in real-to-simulation transfer since one recording can appear in several datasets. For example, one benchmark could reuse the video clips from another benchmark.
The pipeline therefore compares the frames of every training recording with those of the evaluation tasks, and it filters out the recordings that appear in the evaluation tasks, including copies that are re-encoded, resized, or clipped.
For each remaining recording, a method first writes a program, and the pipeline then computes the reward of this program with the same code that \lab{} uses for evaluation.
\lab{} does not train models itself. Instead, the resulting programs and rewards can serve both offline fine-tuning and online reinforcement learning.

So far, \lab{} supports 47 coding agents and 5 non-agent methods, and it adapts 6 existing benchmarks.
All 47 agents, such as OpenHands and Aider, connect through a bridge to the Harbor framework~\cite{harbor2026}, and 5 widely used ones, such as Codex, Claude Code, and Gemini CLI, can also run directly in the sandboxes of \lab.
The 6 benchmarks are Physics-IQ, Video2World, 4DCodeBench, PhysCodeGym, PhySimCode, and IRIS, to which we add a synthetic suite with exact ground truth, and we check each benchmark-wise adapter against the original release of its benchmark.
This paper focuses on the design of the platform, and we leave a systematic comparison of the connected methods to future work.

Last but not least, while positioning a new work, it is challenging to efficiently choose baseline methods and benchmarks. This is because the literature of real-to-simulation transfer scatters over multiple fields, including computer vision, graphics, and robotics.
These papers deliver the results in various ways and styles, from physical parameters and simulation-ready assets to scene-level simulations, simulation programs, and learned simulators.
We therefore also release \textbf{Awesome-Real2Sim}, a research map of nearly 300 real-to-simulation papers (Figure~\ref{fig:overview}, panel~3).
It organizes the papers by motivations or deliverables, and we will keep it up to date as new work appears.

Both the platform and the research map are open source, and we hope that they will grow into a shared effort of \textbf{Real2SimHub}, an open community that we started around both projects, with a website linked under the paper title.
Researchers can connect their own methods, benchmarks, and datasets through the same adapters that we develop, and add their papers to the research map.

Our contributions are as follows:
\begin{itemize}[leftmargin=20pt, itemsep=1pt, topsep=2pt]
  \item \textbf{A unified evaluation platform.}
  \lab{} considers the program that a method delivers as the common interface between methods and benchmarks. As a result, $N$ methods and $M$ benchmarks need only $N + M$ integrations.
  \item \textbf{A pipeline that can create training data and avoid contamination.}
  \lab{} creates training data by filtering the video recordings that appear in any evaluation task, and provides the reward scores for future fine-tuning.
  \item \textbf{An open-source ecosystem that is easy to use and extend.}
  So far, \lab{} supports 47 coding agents and 5 non-agent methods on 6 existing benchmarks.
  Users can combine them in a short experiment file and add new methods or benchmarks as plugins.
  \item \textbf{An up-to-date research map.}
  Awesome-Real2Sim organizes nearly 300 real-to-simulation papers by motivation or deliverable, and we will keep extending it as new work appears.
\end{itemize}

\section{Related Work}
\label{sec:related}
\begin{table}[t]
  \centering
  \caption{\textbf{\lab{} among the evaluation platforms of other fields.}
  Each platform contributes to a research field by offering a shared way to evaluate many methods or systems.
  \lab{} aims to play this role for real-to-simulation, where the method to be tested delivers a simulation program.}
  \label{tab:positioning}
  \footnotesize
  \setlength{\tabcolsep}{4pt}
  \renewcommand{\arraystretch}{1.2}
  \begin{tabularx}{\linewidth}{>{\raggedright\arraybackslash}p{2.6cm}>{\raggedright\arraybackslash}p{2.9cm}Y}
    \toprule
    \textbf{Platform} & \textbf{System under test} & \textbf{Characteristics} \\
    \midrule
    lm-evaluation-harness~\cite{biderman2024lessons} & Language models & Runs many models on many tasks; scores generated text or log-likelihoods against references. \\
    HELM~\cite{liang2022holistic} & Language models & Scores text completions with several metrics in each scenario. \\
    OpenCompass~\cite{opencompass2023} & LLMs and VLMs & Covers language and vision tasks; scores answers by rules or with an LLM judge. \\
    \midrule
    SWE-bench~\cite{jimenez2023swebench} & Coding agents & Checks each patch with unit tests in a container; also releases a training split. \\
    Harbor~\cite{harbor2026} & Agents & Checks the final state of each container with a task-specific verifier; also supports RL rollouts. \\
    \midrule
    SIMPLER~\cite{li2024simpler} & Robot policies & Queries policies for actions step by step; measures task success in simulated versions of real setups. \\
    RoboArena~\cite{atreya2025roboarena} & Generalist robot policies & Runs policies on real robots; ranks them by pairwise preferences of evaluators. \\
    XPolicyLab~\cite{community2026xpolicylab} & Robot policies & Connects $N$ policies to $M$ environments with $N + M$ integrations; also supports real-robot deployment. \\
    \midrule
    \rowcolor{r2sindigo!7}
    \textbf{\lab{} (ours)} & \textbf{Coding agents} & \textbf{Runs the simulation program that each agent writes offline and scores its output with the original benchmark scorer; also creates training data with rewards.} \\
    \bottomrule
  \end{tabularx}
\end{table}

\paragraph{Evaluation platforms in other fields.}
Several fields already share an evaluation platform (Table~\ref{tab:positioning}).
For language models, lm-evaluation-harness~\cite{biderman2024lessons} and HELM~\cite{liang2022holistic} run many models on many tasks with shared metrics, and OpenCompass~\cite{opencompass2023} and VLMEvalKit~\cite{duan2024vlmevalkit} extend this practice to vision--language models.
For agents, SWE-bench~\cite{jimenez2023swebench} scores the patches of coding agents by running unit tests in containers, and Harbor~\cite{harbor2026} runs agents on tasks in containers and checks each task with its own verifier.
For robot policies, SIMPLER~\cite{li2024simpler} evaluates policies in simulated versions of real setups, RoboArena~\cite{atreya2025roboarena} ranks policies through pairwise comparisons on real robots, and XPolicyLab~\cite{community2026xpolicylab} connects $N$ policies to $M$ evaluation environments with $N + M$ integrations.
\lab{} aims to play the same role for real-to-simulation transfer, and like XPolicyLab, it connects methods and benchmarks through adapters.
What passes between the two sides is different, however.
A robot policy is a live model that the environment queries step by step, whereas a method in \lab{} delivers a program once, at the end of its work.
Unlike a live policy, a program can be run again after the session in which it was written, so \lab{} fixes how the program is packaged and re-runs it offline before scoring it (Section~\ref{sec:rerun}).
This re-run also distinguishes \lab{} from Harbor, whose verifier checks the container in which the agent worked, whereas \lab{} scores a frozen re-run of the program outside that container, with the original scorer of each benchmark.
The two platforms are nevertheless complementary, since \lab{} runs the agents of Harbor through a bridge and can export its own tasks as Harbor tasks.

\begin{figure}[h]
  \centering
  \includegraphics[width=\linewidth]{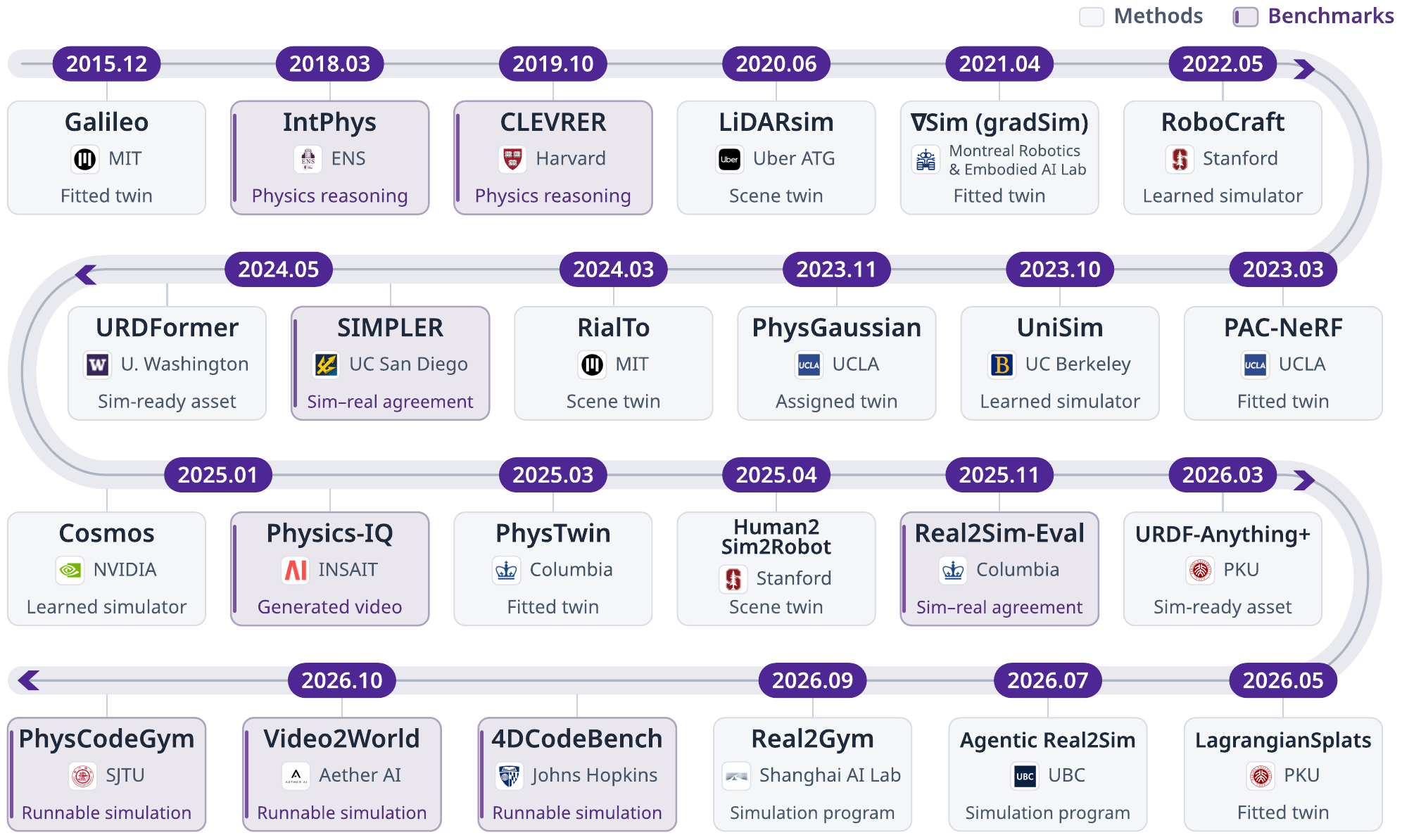}
  \caption{\textbf{A roadmap of real-to-simulation, from Awesome-Real2Sim.}
  Selected methods (gray) and benchmarks (violet) are placed by the month of their first arXiv version (for Galileo, which has none, by the month of its conference).
  Each card names the institution and what the paper delivers or asks for.
  The methods deliver many kinds of results, such as physical twins (fitted or assigned), simulation-ready assets, scene twins, learned simulators, and simulation programs.
  One recent trend is that coding agents write simulation programs, and recent benchmarks such as 4DCodeBench, PhysCodeGym, and Video2World are proposed to evaluate these agents.}
  \label{fig:roadmap}
\end{figure}

\paragraph{A research map of the field.}
Our research map, Awesome-Real2Sim, currently covers nearly 300 real-to-simulation papers, among them more than 150 methods and 50 benchmarks (Figure~\ref{fig:roadmap}).
It assigns each paper to one primary category, according to what the paper delivers or asks for.
Methods fall into five kinds: physical twins, which simulate specific objects with physical parameters that are either fitted to data or assigned by hand or from priors; simulation-ready assets; scene twins, which are scene-level simulations; simulation programs; and learned simulators.
Benchmarks fall into four kinds: those that ask for a runnable simulation, for answers about physics, or for generated frames, and those that test whether a simulator agrees with reality.
The rest of this section uses this map to place \lab{} among methods and benchmarks.

\paragraph{Real-to-simulation methods.}
We highlight three of these kinds, starting with physical twins, which come in two forms.
Fitted twins identify the physical parameters of objects from recordings.
Early works fit a physics engine to videos~\cite{wu2015galileo,wu2017vda}, later works use differentiable simulators~\cite{jatavallabhula2021gradsim,li2023pacnerf}, and PhysTwin~\cite{jiang2025phystwin} extends this approach to deformable objects.
Assigned twins instead assign physical properties to the objects of a reconstructed 3D scene, either by hand or from the priors of generative video models, as in PhysGaussian~\cite{xie2024physgaussian} and PhysDreamer~\cite{zhang2024physdreamer}.
Scene twins rebuild whole environments, such as sensor simulators built from real scans~\cite{manivasagam2020lidarsim} and simulated rooms in which robot policies are trained before they are deployed in the real world~\cite{torne2024rialto,dai2024acdc}.
Simulation programs are the most recent kind: language-model agents write programs that call a physics engine~\cite{cherian2024llmphy,chen2026agentic}, and the newest benchmarks below ask coding agents to write such programs.
\lab{} does not propose a new method; instead, it evaluates methods by the programs that they write, and coding agents, which write such programs directly, are its main focus.

\paragraph{Benchmarks.}
Like methods, benchmarks related to real-to-simulation transfer differ in what they ask a system to deliver.
Benchmarks of physical understanding ask for answers: IntPhys~\cite{riochet2018intphys} asks whether an event is physically possible, CLEVRER~\cite{yi2019clevrer} asks questions about collisions, and Physion~\cite{bear2021physion} asks whether objects will touch.
Other benchmarks score generated videos: Physics-IQ~\cite{motamed2026physics} compares them with real recordings of the same event, while VideoPhy~\cite{bansal2025videophy} judges their physical commonsense.
IRIS~\cite{khanbayov2026iris} instead asks for the physical parameters of recorded systems.
The newest benchmarks ask coding agents to write the simulation itself: a script that builds a 4D world in 4DCodeBench~\cite{shen20264dcodebench}, a Blender scene in PhysCodeGym~\cite{shao2026physcodegym}, a simulation package in Video2World~\cite{tang2026video2world}, and, from simulated videos, a program with its parameters in PhySimCode~\cite{kundu2026physimcode}.
BlenderGym~\cite{gu2025blendergym} also scores Blender programs by their renders, but it targets the editing of 3D scenes rather than the simulation of recordings.
Finally, SIMPLER~\cite{li2024simpler} and RobotArena~$\infty$~\cite{jangir2025robotarena} simulate real robot setups to evaluate robot policies, so they test a use of real-to-simulation transfer rather than the simulation itself.

Rather than defining a new benchmark, \lab{} runs existing benchmarks through one interface and keeps their own scorers whenever they exist.
It adapts the four newest benchmarks above, namely 4DCodeBench, PhysCodeGym, Video2World, and PhySimCode, as well as Physics-IQ and IRIS.
For these two benchmarks, a method delivers a program instead of frames or parameters.
On Physics-IQ, the program renders the frames that the official scorer evaluates.
IRIS, however, scores recovered parameters rather than simulated motion.
\lab{} therefore compares the motion that the program simulates with trajectories extracted from the IRIS videos, so these scores measure reproduction and are not comparable with the parameter-recovery scores of IRIS.
Section~\ref{sec:ecosystem} describes these 6 benchmarks and our synthetic suite.
Two groups of benchmarks remain outside the scope of \lab{}.
Benchmarks that ask only for answers leave no simulation to run, and benchmarks of robot policies, such as SIMPLER, are complementary, since they use simulations built from real setups, whereas \lab{} evaluates how such simulations are built.

\paragraph{Training data and contamination.}
A model may score well on a benchmark simply because its training data contain the test data~\cite{sainz2023nlp}.
Such contamination is hard to detect; for example, rephrased test samples can pass common n-gram decontamination checks~\cite{yang2023rethinking}.
Real-to-simulation transfer faces the same risk in another form: when the test data are recordings, one recording can appear in several datasets after re-encoding, resizing, or clipping.
\lab{} therefore compares the frames of recordings, both when it creates training data and when it evaluates a trained method (Section~\ref{sec:training}).

\section{The \lab{} Platform}
\label{sec:system}

\begin{figure}[b]
  \centering
  \includegraphics[width=\linewidth]{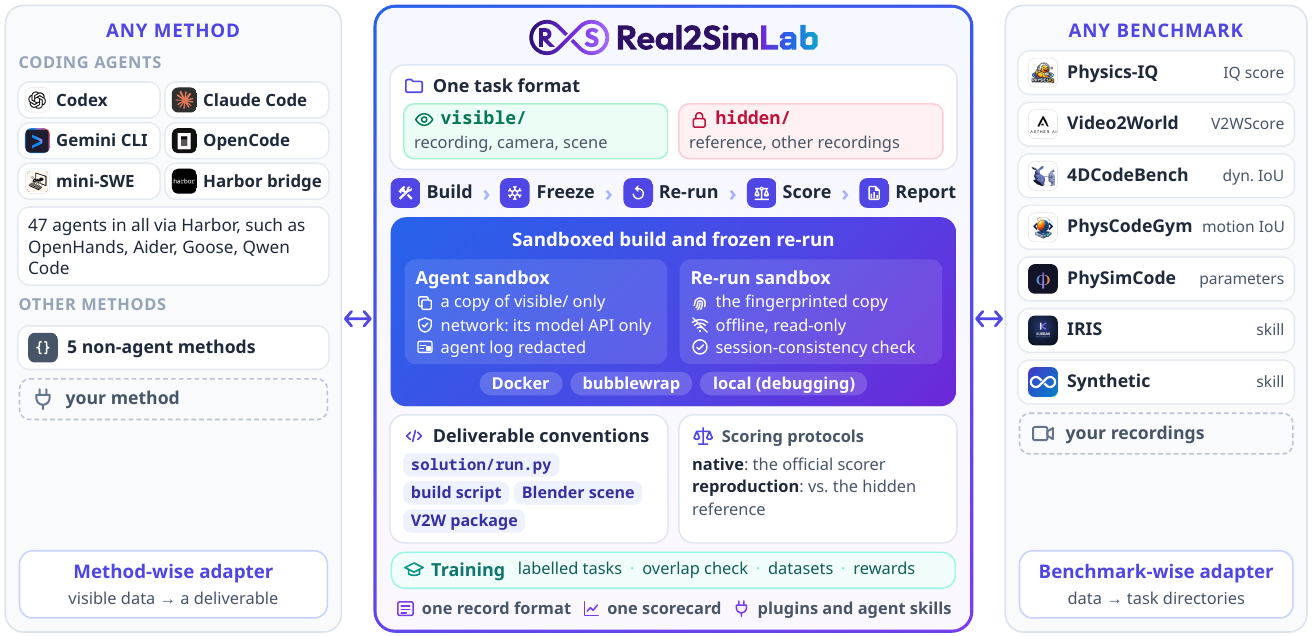}
  \caption{\textbf{Architecture of \lab.}
  A method-wise adapter (left) passes the visible part of a task to a method and collects its deliverable, and a benchmark-wise adapter (right) turns the data of a benchmark into tasks.
  In between (center), \lab{} fixes one task format.
  It builds each deliverable in a sandbox, freezes it, re-runs it offline, and scores it with the protocol of the benchmark.
  All results are stored in one record format and reported in one scorecard. Training tasks use the same format as evaluation tasks, and their rewards are computed by the scoring code of evaluation (Section~\ref{sec:training}).}
  \label{fig:architecture}
\end{figure}

As discussed in Section~\ref{sec:intro}, methods are hard to compare because every benchmark defines its own inputs, deliverables, and scoring pipeline.
To address this, \lab{} sits between methods and benchmarks (Figure~\ref{fig:architecture}).
We call the files that a method hands in for a task its \emph{deliverable}; a deliverable contains a program, and what the program writes when it runs is its \emph{output}.
On each task, a method first \emph{builds} a deliverable from the data that it may see, and \lab{} then \emph{freezes} the deliverable, \emph{re-runs} its program offline, \emph{scores} the output of the re-run, and \emph{reports} the result.
A user describes an evaluation in a short experiment file and runs it with the \cmd{r2s} command-line tool (Listing~\ref{lst:experiment}).
The rest of this section describes the shared task format (Section~\ref{sec:tasks}), how methods build deliverables (Section~\ref{sec:deliveries}), how deliverables are frozen and re-run (Section~\ref{sec:rerun}), and how they are scored and reported (Section~\ref{sec:scoring}).

\subsection{Tasks}
\label{sec:tasks}

A fair evaluation requires that a method never sees the reference that the scorer uses, so each task splits its data into two parts.
Concretely, a task is a directory with a metadata file and two data directories, \cmd{visible/} and \cmd{hidden/}.
The \cmd{visible/} directory holds everything that a method may see, including the recording, the camera parameters when they are known, and a description of the scene.
The \cmd{hidden/} directory holds what only the scorer may see, such as the reference output, the inputs of an official scorer, and other recordings of the same event when they exist (Section~\ref{sec:scoring}).
It is never mounted into the sandbox of a method, and task validation rejects a task whose visible part contains hidden content, names hidden files, or holds symbolic links.

\begin{listing}[tb]
  \begin{minipage}[c]{0.47\linewidth}
\begin{r2slisting}
spec = "r2s-experiment/1.0"
name = "example"
[[tasks.benchmarks]]
name = "physicsiq"
select = ["fill-glass-red-drink/center"]
options = { variants = ["rerun"] }
[[tasks.benchmarks]]
name = "iris"
[[methods]]
name = "codex"
options = { reasoning_effort = "medium" }
[[methods]]
name = "claude_code"
[run]
backend = "docker"
trials = 3
\end{r2slisting}
  \end{minipage}\hfill
  \begin{minipage}[c]{0.5\linewidth}
    \small
\begin{r2slisting}[listing options={basicstyle=\ttfamily\scriptsize, columns=fullflexible}]
$ r2s run example.toml --out runs/example
\end{r2slisting}
    \vspace{0.3em}
    The run directory then contains:
    \begin{itemize}[leftmargin=1.2em, itemsep=0pt, topsep=2pt]
      \item every build, frozen package, re-run, and score;
      \item \cmd{results.json}, which records every result;
      \item a report in HTML, Markdown, and CSV, with the scorecard, its intervals, and reference points.
    \end{itemize}
    \vspace{0.3em}
    To evaluate another connected method or benchmark, a user adds one entry to this file.
  \end{minipage}
  \caption{\textbf{An experiment file.}
  It lists the benchmarks and methods to evaluate and how to run them.
  In this example, two coding agents run three times each, on the center camera view of one Physics-IQ scenario and on all IRIS tasks; for Physics-IQ, only the variant that scores the frozen re-run is selected (Section~\ref{sec:rerun}).}
  \label{lst:experiment}
\end{listing}

Besides the data, the metadata file describes how the task is run: what the method must deliver, namely the convention that the deliverable follows and the output format (Section~\ref{sec:deliveries}), the runtime image, which is the container image with the software that the program may use, the time budget, the hardware, and the license of the data.
From this metadata, \lab{} generates the task statement that a coding agent reads.
When a benchmark written for coding agents provides its own statement, \lab{} uses that statement unchanged.

\subsection{Methods and Deliverables}
\label{sec:deliveries}

Benchmarks ask for different deliverables, and methods produce them in different ways.
To connect each side only once, \lab{} fixes how a method is called and what it may deliver.
A method is anything that turns the visible part of a task into a deliverable, and its method-wise adapter implements a single call, which receives the visible data and returns a deliverable.
For a coding agent, this call starts the agent in a workspace and collects the program that the agent leaves there.

Every deliverable follows one of four conventions, which together cover the 6 benchmarks and our synthetic suite.
Under the convention of \lab{} itself, the deliverable is a program, \cmd{solution/run.py}, that writes its output to a given directory, either as object trajectories (\emph{tracks}) or as a video, together with any values that the benchmark asks for, such as the estimated parameters of PhySimCode.
The other three conventions come from the benchmarks themselves: 4DCodeBench asks for a script that builds a 4D world, PhysCodeGym asks for a Blender scene that renders a video, and Video2World asks for a simulation package that contains the reconstructed scene and the actions of the robot.
Since the task statement tells a coding agent which convention to follow, one agent adapter serves every benchmark, so the $N + M$ integrations of Section~\ref{sec:intro} suffice for coding agents.

Each coding agent works in its own workspace, which holds a copy of the visible data, the task statement, and the documentation of the output format.
The workspace also provides a small self-check tool, which runs the program twice, as the reproducibility check of Section~\ref{sec:rerun} does, and checks that both runs write the same valid output.
Each agent runs in the image that matches the runtime of the task, so it can test its program in the same environment in which the program will later be re-run.
An agent can reach only its model API over the network.
A user may open the network, for example to let an agent install software, but every report then marks the run as such.
Credentials are injected at run time and never written to files, logs, or images, and the logs of agents are redacted before they are stored.

\subsection{Freezing and Re-running}
\label{sec:rerun}

A score should reflect the program alone, not the session in which the program was written.
\lab{} therefore scores the output of a re-run rather than the outputs that a method produced in its session.
To prepare the re-run, it freezes the deliverable, that is, it copies the deliverable and fingerprints the copy.
The frozen program then runs offline, from a read-only directory, without the hidden data, and a re-run refuses any frozen deliverable that no longer matches its fingerprint.
Two sandbox backends enforce these rules: the Docker backend runs each program in a container, and the bubblewrap backend needs neither root access nor a container daemon, so it also works on shared servers and clusters.
A third, local backend runs without isolation for debugging only, and every record that it writes says so.

Re-running also lets \lab{} check each deliverable in two ways.
A deliverable is \emph{session-consistent} if its frozen re-run gives the same output that the method produced in its session, and it is \emph{reproducible} if two re-runs with the same random seed, which \lab{} passes to the program, give the same output.
Both checks compare the full content of the outputs, with a small tolerance for rendering noise in videos, and the scorecard counts how many deliverables pass each check.

Some official protocols, such as those of Physics-IQ and PhysCodeGym, score the output that a method produced in its session.
For such a benchmark, \lab{} builds two variants of each task, so that its scores stay comparable with the official protocol.
One variant scores the output of the session, as the official protocol does, and the other, called \emph{rerun}, scores the output of the frozen re-run.
The two variants are scored as separate tasks, and the scorecard reports both.
\lab{} does not re-run Video2World packages itself.
A Video2World package contains the reconstructed scene and the actions of the robot, and executing it in a simulator requires the official code of Video2World.
Therefore, the official evaluator of Video2World executes the frozen package during scoring.

\subsection{Scoring and Reporting}
\label{sec:scoring}

Each benchmark already defines what a good result is, so \lab{} keeps the original scorer of each benchmark whenever one applies.
Each deliverable is scored with one of two protocols, and the variant of Section~\ref{sec:rerun} decides whether the output of the session or that of the re-run is scored.
The \emph{native} protocol runs the official scorer from a checkout at a pinned commit and stores its report unchanged, so its headline number is the metric that the benchmark itself reports; 5 of the 6 benchmarks use it.
The \emph{reproduction} protocol serves the data for which no official scorer of the program output applies: our synthetic suite, recordings of users, and IRIS, whose own scorer evaluates recovered parameters instead of simulated motion.
It compares the object trajectories of the re-run with the hidden reference.

The headline number of the reproduction protocol is the \emph{reproduction skill}, defined as $1 - e / e_{\text{static}}$.
Here, $e$ is the error of the deliverable under the primary metric that the metadata of the task names, such as the mean distance between the simulated and the recorded positions of the objects, and $e_{\text{static}}$ is the error of a static scene in which every object stays at its first position.
A skill of 1 means that the deliverable reproduces the reference exactly, a skill of 0 means that it does no better than a static scene, and a negative skill means that it does worse.
When the hidden part of a task holds other recordings of the same event, \lab{} also scores each of them against the reference as if it were a deliverable.
Their skill measures how much the real event varies when it is repeated, and a deliverable that exceeds this skill matches the reference more closely than a second recording of the same event does.

A score should also reflect the method rather than the machine.
When the machine fails, for example when a sandbox does not start or an official scorer crashes, \lab{} records an infrastructure error instead of a score, leaves it out of every statistic, and redoes the step the next time the experiment runs.
Every step of an experiment is also keyed by the content of its inputs, so an interrupted experiment resumes where it stopped, and a finished step is reused only while its inputs stay unchanged.

All results share one record format with versioned schemas.
From these records, a scorecard reports the headline number of each benchmark over its tasks, aggregated in the same way as in the original benchmark, or averaged over tasks under the reproduction protocol, with a bootstrap confidence interval over tasks and repeated runs.
Each headline number appears next to reference points on the same tasks, such as the score of a static scene, of another recording of the event, or of the solution released with the benchmark, so that a reader can judge whether a score is high or low.

\section{Ecosystem}
\label{sec:ecosystem}

This section describes the benchmarks and methods that \lab{} connects so far, how we check each benchmark-wise adapter, and how users can extend the platform.

\begin{figure}[t]
  \centering
  \includegraphics[width=\linewidth]{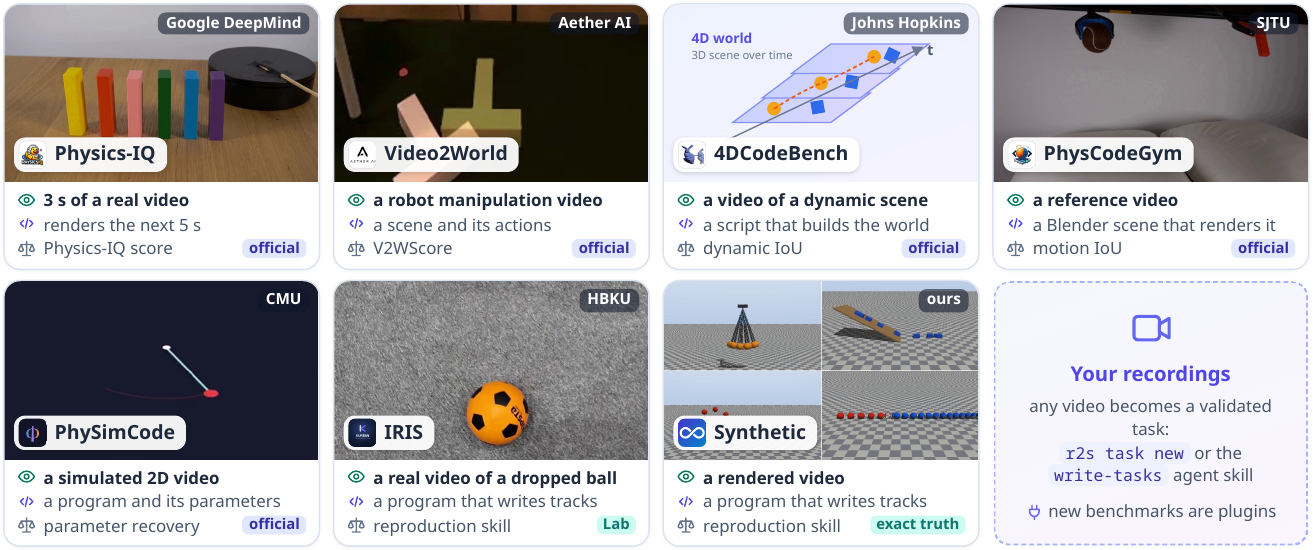}
  \caption{\textbf{The benchmarks that \lab{} adapts so far.}
  Each card shows what a method sees, what it delivers, and how the deliverable is scored.
  Of the 6 existing benchmarks, 5 keep their official scorers, while IRIS (marked \emph{Lab}) and our synthetic suite (marked \emph{exact truth}) use the reproduction skill of \lab.
  New benchmarks and recordings of users are added by converting them into tasks of the shared format (Section~\ref{sec:tasks}).}
  \label{fig:benchmarks}
\end{figure}

\paragraph{Benchmarks.}
So far, \lab{} adapts 6 existing benchmarks and provides a synthetic suite of our own (Figure~\ref{fig:benchmarks}).
Each benchmark-wise adapter downloads the original data at a pinned revision and builds task directories from it.
It also states which official protocol it follows and which parts of the benchmark it leaves out.
Of the 6 benchmarks, 5 keep their official scorers through the native protocol.
IRIS publishes videos of real physical systems but no trajectories, so its adapter extracts the trajectory of the ball from each video and scores deliverables with the reproduction protocol.
The adapter covers the class of IRIS in which a ball is dropped, one of eight classes.
Our synthetic suite renders five phenomena with exact ground truth: a bouncing ball, a damped pendulum, a sliding block, a projectile with drag, and a collision between two balls.
Because its ground truth is exact, the suite serves both as a check of the reproduction protocol and as a source of training tasks (Section~\ref{sec:training}).

\paragraph{Checks against the original releases.}
We check each benchmark-wise adapter against the original release of its benchmark.
For example, the scores that the Physics-IQ adapter reports equal those of the official pipeline on the same videos, and the reference packages of Video2World reproduce their recorded scores on Push-T, a planar pushing task.
On the synthetic suite, the program that generated the data reaches a reproduction skill of 1.
These checks also revealed a few issues in the released data, such as two IRIS recordings that are copies of each other, and the adapters work around and document such issues.

\paragraph{Methods.}
Through a bridge to Harbor 0.27~\cite{harbor2026}, \lab{} connects all 47 agents of Harbor, such as OpenHands, Aider, Cursor CLI, Goose, and Qwen Code.
A bridged agent works in a Harbor container built from the runtime image of the task, with a network that reaches only the model API.
After the agent finishes, \lab{} freezes its program and re-runs it offline before scoring.
In addition, Codex, Claude Code, Gemini CLI, OpenCode, and mini-SWE-agent~\cite{yang2024sweagent} have direct adapters, which run these 5 agents in the sandboxes of \lab{} without Harbor.
Since \lab{} limits the network of every agent and re-runs every program, an agent adapter only needs to state how to call its tool and which credentials to inject.
Besides these agents, \lab{} includes 5 simple non-agent methods, which serve as references and examples.

\paragraph{Extending \lab.}
Benchmarks, methods, metrics, output formats, scorers, sandbox backends, and commands are all plugins.
Their interfaces follow a versioned specification of tasks, deliverables, and records, and within version 1, new releases only add optional fields, so existing plugins keep working.
\lab{} also ships \emph{agent skills}, which are step-by-step procedures that a coding agent can follow to add a benchmark, add a method, or turn a recording into validated tasks.
Through these plugins and agent skills, we hope that the Real2SimHub community will extend \lab{} with new benchmarks, methods, and recordings.

\section{Training with \lab}
\label{sec:training}

\begin{figure}[t]
  \centering
  \includegraphics[width=\linewidth]{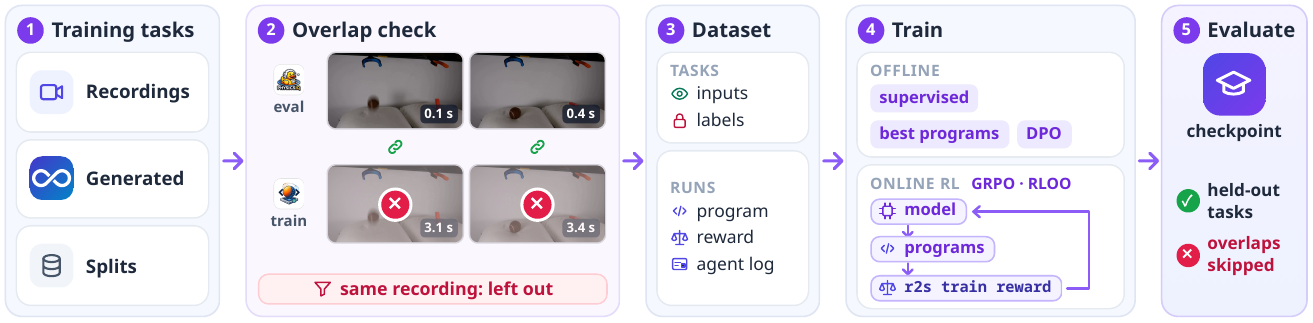}
  \caption{\textbf{The training data pipeline.}
  (1)~Training tasks come from recordings of users, from a released training split when a benchmark has one, or from our generator of synthetic scenes, and they use the format of evaluation tasks.
  (2)~The overlap check removes every task that shows the recording of an evaluation task, by comparing thumbnails of the frames in which something moves, at every time offset.
  In the example, a PhysCodeGym clip, treated here as a candidate for training, contains the hidden test video of a Physics-IQ evaluation task, starting 3\,s into the clip.
  (3)~The dataset holds the tasks and the runs on them, and each run holds a program, its reward, and the log of the agent.
  (4)~A trainer of the user's choice trains a method offline on the dataset, or online with rewards from the evaluation code.
  (5)~\lab{} evaluates the trained method as in Section~\ref{sec:system}, but skips the evaluation tasks that overlap its training tasks.
  \lab{} provides all steps but the trainer; we have not yet trained a model with them.}
  \label{fig:training}
\end{figure}

Besides evaluation, \lab{} provides a pipeline that creates training data with reward annotations for real-to-simulation methods (Figure~\ref{fig:training}).
The pipeline builds training tasks, filters out every task whose recording appears in an evaluation task, and annotates the programs that methods write for the remaining tasks with rewards.
All of these steps follow one rule: each recording serves either training or evaluation, never both.
\lab{} does not ship a trainer or model weights; it provides the data and the rewards, and it keeps a trained method away from the evaluation tasks that overlap its training data.

\paragraph{Training tasks.}
A training task has the format of an evaluation task but is marked for training, and its hidden part holds the labels, such as the reference trajectories, the true parameters, and the true program.
Users can turn their own recordings, together with reference outputs such as object trajectories, into training tasks with the same builder that writes evaluation tasks, and a benchmark-wise adapter can build a training split when a benchmark releases one under terms that allow training.
None of the 6 benchmarks does so: they are released for evaluation, and \lab{} treats them as evaluation sets.
For example, the terms of Physics-IQ forbid using its test data to improve a score, so \lab{} never exports Physics-IQ tasks for training.
To provide training data nonetheless, \lab{} can generate any number of training scenes of the five synthetic phenomena, with exact labels and the true program, and these scenes are the main source of training data today.
Each scene is rendered anew from parameters drawn from fixed ranges, and to avoid near copies of an evaluation scene, a draw is kept only if at least one parameter differs by more than 5\% from that of the evaluation task of the same phenomenon.

\paragraph{Filtering out evaluation recordings.}
Before writing a dataset, \lab{} compares every training task with the evaluation tasks and leaves out each task that overlaps one of them, which guards against data contamination.
Two tasks overlap if they share a file or show the same recording.
To detect the same recording, \lab{} reduces the frames in which something moves to small thumbnails and compares them at every time offset, so that it also catches copies that are re-encoded, resized, clipped, brightened, or reduced to a lower frame rate.
In our tests, the check found 36 of 40 such copies of the synthetic evaluation videos, missing only the most heavily compressed ones, and near its threshold, it errs toward leaving a task out.
On real data, it found each of the 27 clips that PhysCodeGym reuses from Physics-IQ, which shows that a dataset assembled from public recordings can indeed contain the test videos of a benchmark.

\paragraph{Programs and rewards.}
For each training task, a method, such as a coding agent, writes a program in the same way as in an evaluation, and \lab{} annotates the program with a reward.
The reward is the score that the evaluation of the task would give the program, computed by exactly the same code: \lab{} freezes the program, runs it offline, and scores its output.
Since the training tasks are synthetic scenes or recordings of users, this score is the reproduction skill of Section~\ref{sec:scoring}, and the official scorers of the 6 benchmarks, whose data are never used for training, serve evaluation only.
A failure outside the program, such as a sandbox that does not start, gives no reward rather than a reward of 0, so a trainer can tell failures of the method apart from failures of the platform.
\lab{} then exports the training tasks and these runs as one dataset, in which each run holds the program, its reward, and the redacted log of the agent.

\paragraph{Using the data.}
The rewards stored in a dataset serve offline training, such as fine-tuning on high-reward programs and preference learning~\cite{rafailov2023direct} on pairs of programs for the same task.
For online reinforcement learning, each new program can be scored as soon as it is generated, directly or through a Harbor task whose verifier computes the same reward, and since each reward covers a whole program, group-relative methods such as GRPO~\cite{shao2024deepseekmath} and RLOO~\cite{ahmadian2024back} suit this setting.
We have not yet trained a model in these ways.

\paragraph{Evaluating a trained method.}
A dataset is checked only against the evaluation tasks given when it is exported, so the overlap check runs again when a trained method is evaluated.
When a method is trained on an exported dataset through \lab, the thumbnails of its training tasks are stored with the trained model, and in every later evaluation, \lab{} skips each task that overlaps one of them and marks it as not held out, so the score covers held-out tasks only.

\section{Conclusion}
\label{sec:conclusion}

We have presented \lab, a unified platform for real-to-simulation transfer.
\lab{} runs the program that a method writes and evaluates the output simulation with the original scorer of each benchmark.
As a result, the program becomes the common interface between methods and benchmarks.
Evaluating $N$ methods on $M$ benchmarks then requires only $N + M$ integrations instead of $N \times M$.
Moreover, \lab{} creates training data by filtering out the video recordings that appear in any evaluation task, and provides the reward scores for future fine-tuning.
So far, \lab{} supports 47 coding agents and 5 non-agent methods on 6 benchmarks.
We also release Awesome-Real2Sim, a research map of nearly 300 real-to-simulation papers, to help researchers choose baseline methods and benchmarks when positioning a new work.
Both the platform and the research map are open source, and we hope that they will grow into a shared effort of the Real2SimHub community.

\section{Acknowledgments}
\label{sec:acknowledgments}

We thank Haorui Chen, Chenhang Cui, Jinfa Huang, Zheng Lu, Zitong Shen, Fangyuan Shi, Qiwei Wu, Zeyu Xia, Mengyu Yang, and Bofan Zhu (ordered alphabetically by last name) for their valuable comments, insightful discussions, and contributions to this work.

\renewcommand{\bibfont}{\small}
\setlength{\bibsep}{3pt plus 1pt}
\bibliographystyle{plainnat}
\bibliography{references}

\begin{thebibliography}{41}
\providecommand{\natexlab}[1]{#1}
\providecommand{\url}[1]{\texttt{#1}}
\expandafter\ifx\csname urlstyle\endcsname\relax
  \providecommand{\doi}[1]{doi: #1}\else
  \providecommand{\doi}{doi: \begingroup \urlstyle{rm}\Url}\fi

\bibitem[Ahmadian et~al.(2024)Ahmadian, Cremer, Gall{\'e}, Fadaee, Kreutzer,
  Pietquin, {\"U}st{\"u}n, and Hooker]{ahmadian2024back}
Arash Ahmadian, Chris Cremer, Matthias Gall{\'e}, Marzieh Fadaee, Julia
  Kreutzer, Olivier Pietquin, Ahmet {\"U}st{\"u}n, and Sara Hooker.
\newblock Back to basics: Revisiting {REINFORCE} style optimization for
  learning from human feedback in {LLMs}.
\newblock In \emph{Proceedings of the 62nd Annual Meeting of the Association
  for Computational Linguistics (Volume 1: Long Papers)}, pages 12248--12267,
  2024.

\bibitem[{Anthropic}(2026)]{anthropic2026fable}
{Anthropic}.
\newblock Introducing {Claude Fable 5.2} and {Claude Mythos 5.2}.
\newblock Blog post,
  \url{https://www.anthropic.com/claude-fable-and-mythos-5-2}, 2026.

\bibitem[Atreya et~al.(2025)Atreya, Pertsch, Lee, Kim, Jain, Kuramshin, Neary,
  Hu, Arora, Ellis, et~al.]{atreya2025roboarena}
Pranav Atreya, Karl Pertsch, Tony Lee, Moo~Jin Kim, Arhan Jain, Artur
  Kuramshin, Cyrus Neary, Edward~S. Hu, Kanav Arora, Kirsty Ellis, et~al.
\newblock {RoboArena}: Distributed real-world evaluation of generalist robot
  policies.
\newblock In \emph{Conference on Robot Learning}, pages 336--364, 2025.

\bibitem[Bansal et~al.(2025)Bansal, Lin, Xie, Zong, Yarom, Bitton, Jiang, Sun,
  Chang, and Grover]{bansal2025videophy}
Hritik Bansal, Zongyu Lin, Tianyi Xie, Zeshun Zong, Michal Yarom, Yonatan
  Bitton, Chenfanfu Jiang, Yizhou Sun, Kai-Wei Chang, and Aditya Grover.
\newblock {VideoPhy}: Evaluating physical commonsense for video generation.
\newblock In \emph{International Conference on Learning Representations}, 2025.

\bibitem[Bear et~al.(2021)Bear, Wang, Mrowca, Binder, Tung, Pramod, Holdaway,
  Tao, Smith, Sun, Fei-Fei, Kanwisher, Tenenbaum, Yamins, and
  Fan]{bear2021physion}
Daniel~M. Bear, Elias Wang, Damian Mrowca, Felix~J. Binder, Hsiao-Yu~Fish Tung,
  R.~T. Pramod, Cameron Holdaway, Sirui Tao, Kevin Smith, Fan-Yun Sun,
  Li~Fei-Fei, Nancy Kanwisher, Joshua~B. Tenenbaum, Daniel L.~K. Yamins, and
  Judith~E. Fan.
\newblock Physion: Evaluating physical prediction from vision in humans and
  machines.
\newblock In \emph{Proceedings of the Neural Information Processing Systems
  Track on Datasets and Benchmarks}, 2021.

\bibitem[Biderman et~al.(2024)Biderman, Schoelkopf, Sutawika, Gao, Tow, Abbasi,
  Aji, Ammanamanchi, Black, Clive, et~al.]{biderman2024lessons}
Stella Biderman, Hailey Schoelkopf, Lintang Sutawika, Leo Gao, Jonathan Tow,
  Baber Abbasi, Alham~Fikri Aji, Pawan~Sasanka Ammanamanchi, Sidney Black,
  Jordan Clive, et~al.
\newblock Lessons from the trenches on reproducible evaluation of language
  models.
\newblock \emph{arXiv preprint arXiv:2405.14782}, 2024.

\bibitem[Chen et~al.(2026)Chen, Xia, Peng, Zhang, Jing, Ma, Qian, Cheng, Jiao,
  Zhou, et~al.]{chen2026agentic}
Guanxiong Chen, Qianjun Xia, Jiawei Peng, Heng Zhang, Pengyu Jing, Bole Ma,
  Justin Qian, Yixian Cheng, Ziyi Jiao, Bingyang Zhou, et~al.
\newblock {Agentic Real2Sim}: Physics-based world modeling with vision-language
  agents.
\newblock \emph{arXiv preprint arXiv:2607.19190}, 2026.

\bibitem[Cherian et~al.(2026)Cherian, Corcodel, Jain, and
  Romeres]{cherian2024llmphy}
Anoop Cherian, Radu Corcodel, Siddarth Jain, and Diego Romeres.
\newblock {LLMPhy}: Parameter-identifiable physical reasoning combining large
  language models and physics engines.
\newblock In \emph{International Conference on Artificial Intelligence and
  Statistics}, 2026.

\bibitem[Community et~al.(2026)Community, Chen, Chen, Nian, Cai, Chen, Lin,
  Liang, Chen, Xiang, Su, et~al.]{community2026xpolicylab}
XPolicyLab Community, Tianxing Chen, Yue Chen, Tian Nian, Zijian Cai, Guangyu
  Chen, Wenwei Lin, Qiwei Liang, Zanxin Chen, Peicheng Xiang, Kailun Su, et~al.
\newblock {XPolicyLab}: A unified standard and open ecosystem for robot policy
  evaluation and deployment.
\newblock \emph{arXiv preprint arXiv:2608.09892}, 2026.

\bibitem[Dai et~al.(2024)Dai, Wong, Jiang, Wang, Gokmen, Zhang, Wu, and
  Fei-Fei]{dai2024acdc}
Tianyuan Dai, Josiah Wong, Yunfan Jiang, Chen Wang, Cem Gokmen, Ruohan Zhang,
  Jiajun Wu, and Li~Fei-Fei.
\newblock Automated creation of digital cousins for robust policy learning.
\newblock In \emph{Conference on Robot Learning}, 2024.

\bibitem[Duan et~al.(2024)Duan, Fang, Yang, Zhao, Ma, Qiao, Li, Liang, Zhu,
  Agarwal, et~al.]{duan2024vlmevalkit}
Haodong Duan, Xinyu Fang, Junming Yang, Xiangyu Zhao, Zerun Ma, Yuxuan Qiao,
  Mo~Li, Tianhao Liang, Lin Zhu, Amit Agarwal, et~al.
\newblock {VLMEvalKit}: An open-source toolkit for evaluating large
  multi-modality models.
\newblock \emph{arXiv preprint arXiv:2407.11691}, 2024.

\bibitem[Gu et~al.(2025)Gu, Huang, Je, Yang, and Guibas]{gu2025blendergym}
Yunqi Gu, Ian Huang, Jihyeon Je, Guandao Yang, and Leonidas Guibas.
\newblock {BlenderGym}: Benchmarking foundational model systems for graphics
  editing.
\newblock In \emph{Proceedings of the IEEE/CVF Conference on Computer Vision
  and Pattern Recognition}, 2025.

\bibitem[{Harbor Framework Team}(2026)]{harbor2026}
{Harbor Framework Team}.
\newblock {Harbor}: A framework for evaluating and optimizing agents and models
  in container environments.
\newblock Software, \url{https://doi.org/10.5281/zenodo.20953922}, 2026.

\bibitem[Jangir et~al.(2026)Jangir, Zhang, Lo, Yamazaki, Zhang, Tu, Ke, Ke,
  Bisk, and Fragkiadaki]{jangir2025robotarena}
Yash Jangir, Yidi Zhang, Pang-Chi Lo, Kashu Yamazaki, Chenyu Zhang, Kuan-Hsun
  Tu, Tsung-Wei Ke, Lei Ke, Yonatan Bisk, and Katerina Fragkiadaki.
\newblock {RobotArena} $\infty$: Scalable robot benchmarking via real-to-sim
  translation.
\newblock In \emph{International Conference on Learning Representations}, 2026.

\bibitem[Jatavallabhula et~al.(2021)Jatavallabhula, Macklin, Golemo, Voleti,
  Petrini, Weiss, Considine, Parent-Levesque, Xie, Erleben,
  et~al.]{jatavallabhula2021gradsim}
Krishna~Murthy Jatavallabhula, Miles Macklin, Florian Golemo, Vikram Voleti,
  Linda Petrini, Martin Weiss, Breandan Considine, Jerome Parent-Levesque,
  Kevin Xie, Kenny Erleben, et~al.
\newblock {gradSim}: Differentiable simulation for system identification and
  visuomotor control.
\newblock In \emph{International Conference on Learning Representations}, 2021.

\bibitem[Jiang et~al.(2025)Jiang, Hsu, Zhang, Yu, Wang, and
  Li]{jiang2025phystwin}
Hanxiao Jiang, Hao-Yu Hsu, Kaifeng Zhang, Hsin-Ni Yu, Shenlong Wang, and Yunzhu
  Li.
\newblock {PhysTwin}: Physics-informed reconstruction and simulation of
  deformable objects from videos.
\newblock In \emph{Proceedings of the IEEE/CVF International Conference on
  Computer Vision}, pages 7219--7230, 2025.

\bibitem[Jimenez et~al.(2024)Jimenez, Yang, Wettig, Yao, Pei, Press, and
  Narasimhan]{jimenez2023swebench}
Carlos~E. Jimenez, John Yang, Alexander Wettig, Shunyu Yao, Kexin Pei, Ofir
  Press, and Karthik Narasimhan.
\newblock {SWE}-bench: Can language models resolve real-world {GitHub} issues?
\newblock In \emph{International Conference on Learning Representations}, 2024.

\bibitem[Khanbayov et~al.(2026)Khanbayov, Barhdadi, Serpedin, and
  Kurban]{khanbayov2026iris}
Rasul Khanbayov, Mohamed~Rayan Barhdadi, Erchin Serpedin, and Hasan Kurban.
\newblock {IRIS}: A real-world benchmark for inverse recovery and
  identification of physical dynamic systems from monocular video.
\newblock In \emph{European Conference on Computer Vision}, pages 366--383.
  Springer, 2026.

\bibitem[Kundu et~al.(2026)Kundu, Gupta, Julin, Zhao, Xie, and
  Jeni]{kundu2026physimcode}
Souraja Kundu, Aditya Gupta, Joel Julin, Yizhou Zhao, Liuyue Xie, and Laszlo~A.
  Jeni.
\newblock {PhySimCode}: A benchmark and evaluation method for physics video to
  code generation.
\newblock \emph{arXiv preprint arXiv:2610.15722}, 2026.

\bibitem[Li et~al.(2023)Li, Qiao, Chen, Jatavallabhula, Lin, Jiang, and
  Gan]{li2023pacnerf}
Xuan Li, Yi-Ling Qiao, Peter~Yichen Chen, Krishna~Murthy Jatavallabhula, Ming
  Lin, Chenfanfu Jiang, and Chuang Gan.
\newblock {PAC-NeRF}: Physics augmented continuum neural radiance fields for
  geometry-agnostic system identification.
\newblock In \emph{International Conference on Learning Representations}, 2023.

\bibitem[Li et~al.(2024)Li, Hsu, Gu, Pertsch, Mees, Walke, Fu, Lunawat, Sieh,
  Kirmani, et~al.]{li2024simpler}
Xuanlin Li, Kyle Hsu, Jiayuan Gu, Karl Pertsch, Oier Mees, Homer~Rich Walke,
  Chuyuan Fu, Ishikaa Lunawat, Isabel Sieh, Sean Kirmani, et~al.
\newblock Evaluating real-world robot manipulation policies in simulation.
\newblock In \emph{Conference on Robot Learning}, 2024.

\bibitem[Liang et~al.(2023)Liang, Bommasani, Lee, Tsipras,
  et~al.]{liang2022holistic}
Percy Liang, Rishi Bommasani, Tony Lee, Dimitris Tsipras, et~al.
\newblock Holistic evaluation of language models.
\newblock \emph{Transactions on Machine Learning Research}, 2023.

\bibitem[Manivasagam et~al.(2020)Manivasagam, Wang, Wong, Zeng, Sazanovich,
  Tan, Yang, Ma, and Urtasun]{manivasagam2020lidarsim}
Sivabalan Manivasagam, Shenlong Wang, Kelvin Wong, Wenyuan Zeng, Mikita
  Sazanovich, Shuhan Tan, Bin Yang, Wei-Chiu Ma, and Raquel Urtasun.
\newblock {LiDARsim}: Realistic {LiDAR} simulation by leveraging the real
  world.
\newblock In \emph{Proceedings of the IEEE/CVF Conference on Computer Vision
  and Pattern Recognition}, pages 11167--11176, 2020.

\bibitem[Motamed et~al.(2026)Motamed, Culp, Swersky, Jaini, and
  Geirhos]{motamed2026physics}
Saman Motamed, Laura Culp, Kevin Swersky, Priyank Jaini, and Robert Geirhos.
\newblock Do generative video models understand physical principles?
\newblock In \emph{Proceedings of the IEEE/CVF Winter Conference on
  Applications of Computer Vision}, pages 948--958, 2026.

\bibitem[{OpenAI}(2026)]{openai2026gpt6}
{OpenAI}.
\newblock {GPT-6 Astra}: A new generation of intelligence.
\newblock Blog post, \url{https://openai.com/index/gpt-6-astra/}, 2026.

\bibitem[{OpenCompass Contributors}(2023)]{opencompass2023}
{OpenCompass Contributors}.
\newblock {OpenCompass}: A universal evaluation platform for foundation models.
\newblock Software, \url{https://github.com/open-compass/opencompass}, 2023.

\bibitem[Rafailov et~al.(2023)Rafailov, Sharma, Mitchell, Ermon, Manning, and
  Finn]{rafailov2023direct}
Rafael Rafailov, Archit Sharma, Eric Mitchell, Stefano Ermon, Christopher~D.
  Manning, and Chelsea Finn.
\newblock Direct preference optimization: Your language model is secretly a
  reward model.
\newblock In \emph{Advances in Neural Information Processing Systems},
  volume~36, pages 53728--53741, 2023.

\bibitem[Riochet et~al.(2018)Riochet, Castro, Bernard, Lerer, Fergus, Izard,
  and Dupoux]{riochet2018intphys}
Ronan Riochet, Mario~Ynocente Castro, Mathieu Bernard, Adam Lerer, Rob Fergus,
  V{\'e}ronique Izard, and Emmanuel Dupoux.
\newblock {IntPhys}: A framework and benchmark for visual intuitive physics
  reasoning.
\newblock \emph{arXiv preprint arXiv:1803.07616}, 2018.

\bibitem[Sainz et~al.(2023)Sainz, Campos, Garc{\'i}a-Ferrero, Etxaniz,
  de~Lacalle, and Agirre]{sainz2023nlp}
Oscar Sainz, Jon~Ander Campos, Iker Garc{\'i}a-Ferrero, Julen Etxaniz,
  Oier~Lopez de~Lacalle, and Eneko Agirre.
\newblock {NLP} evaluation in trouble: On the need to measure {LLM} data
  contamination for each benchmark.
\newblock In \emph{Findings of the Association for Computational Linguistics:
  EMNLP 2023}, pages 10776--10787, 2023.

\bibitem[Shao et~al.(2026)Shao, He, Fan, Guo, Wang, Ren, Zhang, Xu, Li, Wang,
  Lin, Liu, and Zhang]{shao2026physcodegym}
Shuai Shao, Jinchen He, Cunxin Fan, Zihan Guo, Shijian Wang, Qihan Ren,
  Kangning Zhang, Zunnan Xu, Haoyuan Li, Hao Wang, Jianghao Lin, Weiwen Liu,
  and Weinan Zhang.
\newblock {PhysCodeGym}: Benchmarking coding agents for physically verifiable
  {4D} reconstruction.
\newblock \emph{arXiv preprint arXiv:2610.17288}, 2026.

\bibitem[Shao et~al.(2024)Shao, Wang, Zhu, Xu, Song, Bi, Zhang, Zhang, Li, Wu,
  and Guo]{shao2024deepseekmath}
Zhihong Shao, Peiyi Wang, Qihao Zhu, Runxin Xu, Junxiao Song, Xiao Bi, Haowei
  Zhang, Mingchuan Zhang, Y.~K. Li, Y.~Wu, and Daya Guo.
\newblock {DeepSeekMath}: Pushing the limits of mathematical reasoning in open
  language models.
\newblock \emph{arXiv preprint arXiv:2402.03300}, 2024.

\bibitem[Shen et~al.(2026)Shen, Kova{\v{c}}i{\v{c}}, Kulits, Wang, Li,
  Tenenbaum, Yuille, Chen, and Wu]{shen20264dcodebench}
Ruihong Shen, {\v{Z}}iga Kova{\v{c}}i{\v{c}}, Peter Kulits, Xingrui Wang,
  Zizhang Li, Joshua~B. Tenenbaum, Alan Yuille, Jieneng Chen, and Jiajun Wu.
\newblock {4DCodeBench}: Benchmarking agents on inverse graphics of dynamic
  scenes.
\newblock \emph{arXiv preprint arXiv:2610.03715}, 2026.

\bibitem[Tang et~al.(2026)Tang, Zhang, Yang, Yan, Zhou, Mao, Han, Cao, Xu, He,
  Fu, Feng, and Huang]{tang2026video2world}
Jinzhou Tang, Zijun Zhang, Jing Yang, Yuchen Yan, Kun Zhou, Lingjun Mao,
  Ruobing Han, Jinglin Cao, Wenpeng Xu, Lukun He, Minghao Fu, Fan Feng, and
  Biwei Huang.
\newblock {Video2World}: Benchmarking coding agents for interactive world
  modeling from embodied videos.
\newblock \emph{arXiv preprint arXiv:2610.04432}, 2026.

\bibitem[Torne et~al.(2024)Torne, Simeonov, Li, Chan, Chen, Gupta, and
  Agrawal]{torne2024rialto}
Marcel Torne, Anthony Simeonov, Zechu Li, April Chan, Tao Chen, Abhishek Gupta,
  and Pulkit Agrawal.
\newblock Reconciling reality through simulation: A real-to-sim-to-real
  approach for robust manipulation.
\newblock In \emph{Robotics: Science and Systems}, 2024.

\bibitem[Wu et~al.(2015)Wu, Yildirim, Lim, Freeman, and
  Tenenbaum]{wu2015galileo}
Jiajun Wu, Ilker Yildirim, Joseph~J. Lim, William~T. Freeman, and Joshua~B.
  Tenenbaum.
\newblock Galileo: Perceiving physical object properties by integrating a
  physics engine with deep learning.
\newblock In \emph{Advances in Neural Information Processing Systems},
  volume~28, 2015.

\bibitem[Wu et~al.(2017)Wu, Lu, Kohli, Freeman, and Tenenbaum]{wu2017vda}
Jiajun Wu, Erika Lu, Pushmeet Kohli, William~T. Freeman, and Joshua~B.
  Tenenbaum.
\newblock Learning to see physics via visual de-animation.
\newblock In \emph{Advances in Neural Information Processing Systems}, pages
  153--164, 2017.

\bibitem[Xie et~al.(2024)Xie, Zong, Qiu, Li, Feng, Yang, and
  Jiang]{xie2024physgaussian}
Tianyi Xie, Zeshun Zong, Yuxing Qiu, Xuan Li, Yutao Feng, Yin Yang, and
  Chenfanfu Jiang.
\newblock {PhysGaussian}: Physics-integrated {3D} gaussians for generative
  dynamics.
\newblock In \emph{Proceedings of the IEEE/CVF Conference on Computer Vision
  and Pattern Recognition}, 2024.

\bibitem[Yang et~al.(2024)Yang, Jimenez, Wettig, Lieret, Yao, Narasimhan, and
  Press]{yang2024sweagent}
John Yang, Carlos~E. Jimenez, Alexander Wettig, Kilian Lieret, Shunyu Yao,
  Karthik Narasimhan, and Ofir Press.
\newblock {SWE}-agent: Agent-computer interfaces enable automated software
  engineering.
\newblock In \emph{Advances in Neural Information Processing Systems},
  volume~37, pages 50528--50652, 2024.

\bibitem[Yang et~al.(2023)Yang, Chiang, Zheng, Gonzalez, and
  Stoica]{yang2023rethinking}
Shuo Yang, Wei-Lin Chiang, Lianmin Zheng, Joseph~E. Gonzalez, and Ion Stoica.
\newblock Rethinking benchmark and contamination for language models with
  rephrased samples.
\newblock \emph{arXiv preprint arXiv:2311.04850}, 2023.

\bibitem[Yi et~al.(2020)Yi, Gan, Li, Kohli, Wu, Torralba, and
  Tenenbaum]{yi2019clevrer}
Kexin Yi, Chuang Gan, Yunzhu Li, Pushmeet Kohli, Jiajun Wu, Antonio Torralba,
  and Joshua~B. Tenenbaum.
\newblock {CLEVRER}: Collision events for video representation and reasoning.
\newblock In \emph{International Conference on Learning Representations}, 2020.

\bibitem[Zhang et~al.(2024)Zhang, Yu, Wu, Feng, Zheng, Snavely, Wu, and
  Freeman]{zhang2024physdreamer}
Tianyuan Zhang, Hong-Xing Yu, Rundi Wu, Brandon~Y. Feng, Changxi Zheng, Noah
  Snavely, Jiajun Wu, and William~T. Freeman.
\newblock {PhysDreamer}: Physics-based interaction with {3D} objects via video
  generation.
\newblock In \emph{European Conference on Computer Vision}, 2024.

\end{thebibliography}

\end{document}